\documentclass[pdflatex,sn-mathphys-num]{sn-jnl}

\usepackage{array}
\usepackage{longtable}
\usepackage{graphicx}%
\usepackage{multirow}%
\usepackage{amsmath,amssymb,amsfonts}%
\usepackage{amsthm}%
\usepackage{mathrsfs}%
\usepackage[title]{appendix}%
\usepackage{xcolor}%
\usepackage{textcomp}%
\usepackage{manyfoot}%
\usepackage{booktabs}%
\usepackage{listings}%
\usepackage{pdflscape}
\usepackage{subcaption}
\usepackage{caption}
\usepackage[ruled,vlined]{algorithm2e}

\theoremstyle{thmstyleone}%
\theoremstyle{thmstyletwo}%

\theoremstyle{thmstylethree}%

\begin{document}
\sloppy

\title{Financial Audit Assistance using Misinformation Detection and Explanation}


\author*[1]{\fnm{Kshitij} \sur{Madhav Jadhav}}\email{kshitij.jadhav@tcs.com}

\author[1]{\fnm{Sushodhan} \sur{Vaishampayan}}\email{sushodhan.sv@tcs.com}

\author[1]{\fnm{Manoj} \sur{Apte}}\email{manoj.apte@tcs.com}

\author[1]{\fnm{Sachin} \sur{Pawar}}\email{sachin7.p@tcs.com}

\author[1]{\fnm{Nitin} \sur{Ramrakhiyani}}\email{nitin.ramrakhiyani@tcs.com}

\author[2]{\fnm{Girish} \sur{Keshav Palshikar}}\email{girishpalshikar@gmail.com}

\affil[1]{\orgname{TCS Research}, \orgaddress{\city{Pune}, \postcode{411057}, \state{Maharashtra}, \country{India}}}

\affil[2]{Work done while working at TCS Research}

\abstract{Financial statements (FS) such as Balance Sheet (BS), Income Statement (IS) and Cash-flow Statement (CS) summarize the annual financial performance of a company. FS are widely used for evaluating corporate governance, credit appraisal, risk analysis, validate taxation, make investment decisions etc. Financial auditing is a complex and knowledge-intensive discipline whose one important aim is ensuring integrity, accuracy, fairness and absence of material misstatement in the published FS. Given the importance of FS, there are incentives to hide, omit or falsify information to misrepresent the true financial health of the company; e.g., reduce tax liabilities, or increase investor confidence. Given the complex, time-consuming and expertise-dependent nature of auditing, auditors would benefit from an AI-assisted system that automatically detects instances of misinformation in the given FS and identify likely sources of this misinformation in the financial data. In this paper, we present unsupervised techniques to identify misinformation in FS, and also generate explanations as to the financial variables that are likely sources of misinformation. The auditor can then explore in more detail the associated data sources and business processes to validate these suggestions. A crucial feature of our approach is the use of past corpus of FS and associated audit reports to generate insights, which help in providing assistance. We demonstrate the efficacy of these techniques on a large corpus of 11,460 FS over 5 years and associated audit reports. This paper integrates and adds more novel contributions over the previously reported research (Shinde et al., 2022)\cite{SVAP22}, (Vaishampayan et al., 2022)\cite{VSPP22}, (Pawar et al., 2023)\cite{PAPV23}, which we have used as the foundation for our AI-assisted Auditor Assistance system. }




\maketitle

\section{Introduction}

{\em Financial auditing} is a complex and knowledge-intensive discipline within accounting, whose important aims include ensuring integrity of published financial information, building trust in the business processes and practices, and ensuring quality of governance in an organization~\cite{AEB16}. A {\em financial audit} (or just {\em audit}) consists of many inter-related tasks such as: validation of internal controls, safeguarding the assets, evaluation of current and future risks, providing suggestions for improving the governance, ensuring that processes are followed as required, ensuring compliance with standards, guidelines, laws and regulations and ensuring that the reported financial information is fair and accurate. Stakeholders use audited {\em financial statements} (FS), such as {\em balance sheet} (BS), income statement, cash-flow statement etc., for decision-making. Examples: regulatory bodies use FS to check compliance, tax departments use FS to validate taxes paid and benefits claimed, investors use FS to estimate the financial health of the company and to make investment decisions. 

A {\em financial auditor} is responsible for carrying out financial audit of an organization. In this paper, we focus on the important outcome of an audit viz., validation and certification that the financial data mentioned in the FS of a company is fair and accurate and free of misinformation. Accountants adhere to {\em Generally Accepted Accounting Principles (GAAP)}\footnote{https://asc.fasb.org/} and {\em International Financial Reporting Standards (IFRS)}\footnote{https://www.ifrs.org/} when preparing FS. In many practical domains—such as corporate governance, credit appraisal, risk analysis, taxation, auditing, and investment decisions—the contents of financial statements (which are typically publicly available for listed companies) are thoroughly analyzed from multiple perspectives~\cite{FL07}, \cite{DF12}. Thus validation of the quality of information in FS is an important task in audit. This validation is typically done by an auditor by collecting evidence from (a) trails of business processes followed in the company (e.g., payment receipts, transaction statements, contractual documents, letters to and from banks, authorities, customers and suppliers etc.); as well as from (b) physical inspections (e.g., of warehouses). If the data in FS is consistent with the evidence collected, then the auditor declares in an {\em audit report}  that the FS, are fair and accurate, and presented in accordance with the relevant accounting standards. If not, the auditor makes {\em adverse remarks} about the detected or potential instances of non-conformance, misinformation, irregularities, inconsistencies, errors, inaccuracies, frauds, lapses, non-compliance, violations etc. Adverse remarks often also include auditors' suggestions for improvement. Clearly, knowledge and experience of the auditor play a vital role to efficiently and effectively carry out an audit.

Given the significance of FS, companies often have strong incentives to conceal, omit, or manipulate information to distort their true financial position—for example, to minimize tax liabilities or boost investor confidence.~\cite{BCH99}. Common forms of misinformation in financial statements include inflating the firm’s assets, revenues, or profits, and understating its liabilities, expenses, or losses\footnote{https://www.acfe.com}. 
It is widely recognized that numerous major corporate frauds have their roots in accounting misrepresentation; e.g., Enron in 2001 in USA~\cite{HP03}, WorldCom in 2003 in USA\footnote{https://www.scu.edu/ethicsI-areas/businessethics/
resources/worldcom/} and, Satyam Computers in 2009 in India~\cite{Bhas13}. Misinformation in FS can cause significant monetary losses to customers, employees, and creditors, while also damaging the company’s reputation, trust, and goodwill, if caught. Auditors and forensic accountants employ a variety of investigative techniques to verify the sources of figures reported in financial statements and, in doing so, detect potential misinformation~\cite{Nigr20}. These techniques are largely manual and rely heavily on the deep domain expertise of human professionals. Due to the effort-intensive and subjective nature of such investigations, numerous analytical methods have been developed to identify financial statements that are likely to contain misinformation

\begin{figure}[!ht]
\centering
\small
\includegraphics[width=\columnwidth]{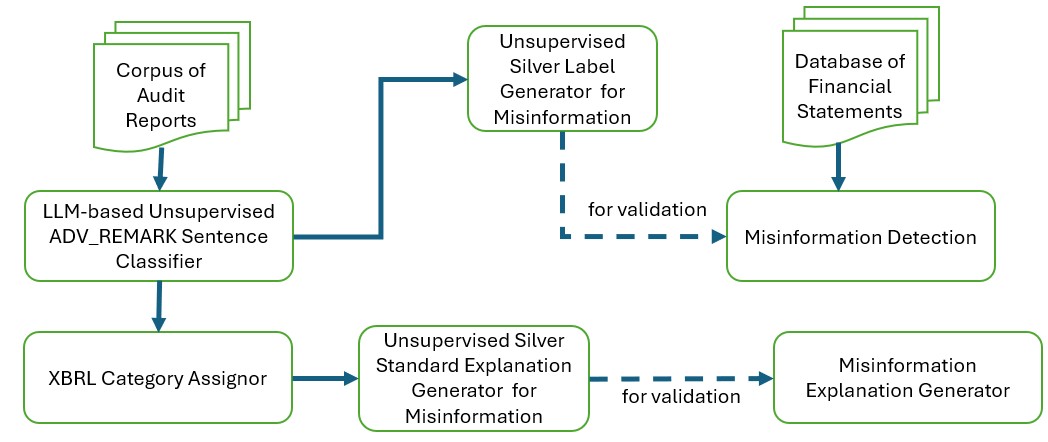}
\caption{Components and overall flow of the audit assistance facilities.}
\label{img:audit_flow1}
\end{figure}

Considering these challenges, in this paper we consider developing techniques to provide assistance to the auditors. Specifically, we consider the scenario where an auditor is conducting financial audit of a particular company for a particular financial year and has access to the associated FS. We propose the following auditor assistance approach: use large external database(s) of semi-structured FS of various companies as well as an associated corpus of textual audit reports; derive insights from these inputs; then use these inputs to provide assistance to the auditor for the following tasks: 
\begin{enumerate}
\item \textbf{Misinformation Detection}: Classify the given FS as likely to contain financial misinformation or not, and validate the results using the silver labels (explained below). 
\item \textbf{Explanation of Misinformation}: If yes, then we develop techniques to generate a human-understandable explanation as to the nature of likely financial misinformation in the FS. 
\item \textbf{Audit Hypotheses Generation}: Generate suggestions (hypotheses) to auditors for more detailed exploration.
\end{enumerate}
As part of pre-processing, we also generate silver labels for each FS in the input database. Since ground-truth labels for FS - whether it contains any instance of misinformation or not - are usually not available, we develop techniques for using the associated audit report to automatically assign a silver label to the FS. Figure~\ref{img:audit_flow1} shows the conceptual flow of the assistance facilities in our system. 

This paper essentially integrates the separate pieces of previously reported research \cite{SVAP22}, \cite{PAPV23}, \cite{VSPP22}, which we have used as the foundation for our AI-assisted Auditor Assistance system. Apart from conceptually integrating previous techniques for auditor assistance, this paper also adds more novel contributions, as follows.  
\begin{enumerate}
    \item The FS data we used in this paper is much bigger as compares to previous works. We use the FS data and audit reports of 2292 companies for five years (11,460 in total), unlike the earlier work where only one year's data was used. Earlier work also used only balance sheets, whereas we integrate multiple FS: balance sheet, income statement (IS), and cash-flow statement (CS).  
    \item In this paper, we extend the number of financial variables in our dataset than those used earlier, by including additional financial variables from Income Statement and Cash-flow Statement. This increases the effectiveness of the techniques for detecting misinformation in the FS.
    \item We propose two different strategies to use audit reports and generative language models to create the silver-labelled data for detecting misinformation in FS. This approach is scalable and can be used to automatically label all FS in our dataset. The datasets in the earlier work used a small manually labeled subset of FS for validation. 
    \item We use a T5\cite{raffel2020exploring} based encoder-decoder model with generative capability for the classification of sentences in audit reports. This model provides the classification label and the explanation for classification. This is an improved model as compared to the \textit{Attn-BERT} model used in the earlier research \cite{PAPV23}.
    \item Unlike the previous work (which used manually generated explanations), we propose an unsupervised method to automatically generate silver standard explanations for FS identified as anomalous i.e., likely to contain misinformation. We use the international standard for financial vocabulary (called XBRL) in performing this task. The idea is to identify adverse remarks in the audit report, assign XBRL categories to each of them, and map these categories to the financial variables we use, which constitutes the silver standard explanation. 
    \item Effectively, our integrated system removes manual labour involved in the various steps in generating assistance, making the system efficient, scalable and more objective. 
\end{enumerate}

A quick summary of motivations for providing AI-based assistance to auditors is as follows. Given the critical importance of audits and the substantial demands on the knowledge, experience, and effort of auditing teams, it is essential to minimize reliance on manual processes and human expertise during audits. Such assistance can enhance both the effectiveness and efficiency of audits by reducing time, cost, and effort. Poor-quality audits—whether intentional or accidental—can lead to severe consequences, including fraud, loss of earnings, erosion of goodwill, legal disputes, inability to continue as a {\em going concern}, and even bankruptcy; e.g., see~\cite{LL19}. There are many reasons why an audit can be of poor quality: lack of expertise in the auditing team~\cite{RW10}, compromised auditor independence~\cite{TL15}, auditor biases, conservatism (recognize bad news rather than good news)~\cite{Bas97} and risk-averse attitudes of auditors, non-cooperation from management, insufficient time/efforts spent in auditing etc. The Sarbanes-Oxley Act of 2002 was specifically introduced to strengthen auditing practices and improve public information disclosure, following a series of persistent scandals driven by auditing failures such as Enron~\cite{beasley1999fraudulent} and Satyam~\cite{Bhas13}. A comprehensive and high-quality audit report serves as a key indicator of an effective audit. Monitoring bodies, such as the Institute of Chartered Accountants (CA) Society of India , have issued guidelines specifying the essential components of audit reports, outlining the aspects auditors should address and describe. As we show in this paper, AI can extract useful insights from past audits and use them to provide effective assistance to improving various auditing tasks in a new audit, and thereby improving the quality of the new audit. 

This paper is organized as follows. In section \ref{sec:related_work} we review some of the related work. Section \ref{sec:datasets} contains the description of the various datasets used in this paper. In this section, we also develop a technique to use pre-trained {\em large language models} (LLMs) for a binary classification of sentences from audit reports into adverse remark or not (e.g., \cite{PAPV23}). A simple technique for generating silver-labels for each FS (whether it contains misinformation or not) is also described in this section. Use of unsupervised regression based techniques and anomaly detection techniques to detect misinformation in structured FS data is described in section \ref{sec:misinfo}. Section \ref{sec:expl} proposes techniques for automatically generating human-understandable explanations for each FS predicted to likely contain instance(s) of misinformation. This explanation takes the form of a set of financial variables responsible for marking the companies come out as anomalous in our predictive techniques for misinformation detection. Section \ref{sec:audit_suggestions} describes the techniques that can be used for generating suggestions to assist the auditor. Finally, we conclude our discussion in section \ref{sec:conc} and add the tasks to be done in the future.

\section{Related Work}\label{sec:related_work}

\textbf{Supervised Detection of Misinformation in FS}: Each FS is typically represented as a vector of well-established financial ratios. In addition to FS, many companies also publish an auditor’s report, which highlights any issues or discrepancies identified during the audit. Financial statements flagged in such reports are referred to as {\em qualified}. A FS with labelled as positive case of misinformation if it is qualified and negative case if it is {\em unqualified}, to create a labelled dataset of FS. \cite{Spa02} trained a binary logistic regression classifier on a labeled dataset comprising FS from 76 manufacturing firms (38 fraudulent and 38 non-fraudulent), each represented by 10 financial ratios. Similarly, \cite{KKTT06} analyzed FS of 164 firms for 2001–02 (41 fraudulent), using 23 financial ratios. They applied rank influence to select 8 key ratios and trained multiple classification models, achieving up to 96.7\% accuracy with a stacking ensemble approach. \cite{Che16} proposed a two-stage framework for detecting fraudulent FS: Stage 1 employed CART and CHAID models to identify significant variables, while Stage 2 combined classifiers such as CART, CHAID, Bayesian Belief Networks, SVMs, and artificial neural networks for fraud detection. A similar methodology was adopted by \cite{Jan18} on 160 companies (40 fraudulent) from the Taiwan Stock Exchange, yielding an accuracy of 90.83\% in detecting fraudulent statements.

\textbf{Unsupervised Detection of Misinformation in FS}:  \cite{MM20} applied Self-Organizing Maps (SOM) to an unlabeled dataset of FS from 1,560 South African municipalities, each represented by three financial ratios. The resulting map was clustered using K-Means, and adverse auditor comments were used to label certain clusters as fraudulent. Similarly, \cite{LTV19} employed a Mahalanobis distance-based anomaly detection technique on an unlabeled dataset of 937 Vietnamese companies (each represented by 24 financial ratios) to identify anomalous FS potentially containing misinformation. Another form of accounting error occurs when semantically equivalent facts reported in different sections of a FS have inconsistent numerical values. To address this, \cite{LYCY20} proposed a neural network-based system for determining whether two table cells refer to the same fact. Beyond FS, misinformation is also prevalent in other financial documents, such as tax reports, primarily for tax evasion purposes. Unsupervised techniques are commonly used to detect suspicious tax reports \cite{dPMd18}, \cite{MdM15}, \cite{GV13}, \cite{WOLC12}. Broadly, these methods involve clustering tax reports, learning patterns characterizing each cluster (e.g., association rules, in-cluster probability distributions, cluster-specific CHAID classification trees), and then using these patterns to identify suspicious reports.

\textbf{Explanations for Anomalies}: A basic explanation for an outlier can be derived by identifying the subspace where the point is most distinct from others. These outlying aspects~\cite{zhang2004hos} are typically determined using two approaches: (i) selecting the top $k$ subspaces with the highest anomaly scores (Score and Search), or (ii) identifying a small, relevant subspace aligned with feature selection objectives~\cite{samariya2020comprehensive}. \cite{zhang2004hos} introduced a distance-based metric called outlying degree (OD) and a dynamic subspace search framework, HOS-miner, to locate subspaces where a query object is an outlier. OAMiner~\cite{duan2015mining} employs a heuristic search strategy, ranking subspaces based on kernel density estimation of the query object. To address the challenge of searching through exponentially many subspaces, \cite{vinh2016discovering} proposed dimensionally unbiased methods such as density Z-score and iPath, combined with a beam search algorithm. OARank~\cite{vinh2015scalable}, a hybrid framework, integrates the efficiency of feature selection with the versatility of score-and-search methods. In its first stage, features are ranked by their potential to make a point outlying; in the second stage, score-and-search is applied to a smaller subset of the top-ranked $k \ll m$ features, where $m$ is the total number of features.

{\em Local Outliers with Graph Projection} (LOGP)~\cite{dang2014discriminative} introduces a set of objective functions to learn the local discriminating subspace for a point using a graph-based transformation. The outlier score is computed as the statistical distance between the point and its neighbors in the transformed subspace. \cite{angiulli2017outlying} proposed a criterion that estimates the probability density function (pdf) of an attribute value for an outlier relative to the pdf of the same attribute across other instances—the lower the probability, the more likely the instance is an outlier. Anomaly Contribution Explainer (ACE)~\cite{zhang2019ace} and its variant ACE-KL provide feature-level contributions as a vector of real numbers. ACE approximates the neighborhood of an outlier by generating synthetic neighbors and fitting a linear regression model with a modified loss function, while ACE-KL adds a regularizer to maximize the KL divergence between a uniform distribution and the computed contribution distribution. \cite{siddiqui2019sequential} introduced Sequential Feature Explanations (SFE), which present features one at a time to users until a confident anomaly judgment can be made, formulated as an optimization problem.

The Explainer~\cite{kopp2020anomaly} provides explanations in the form of rule disjunctions derived from decision trees within a random forest for a given anomalous point. LOOKOUT~\cite{gupta2018beyond}, given a set of outliers and their feature sets, generates an optimal number of 2-D focus plots based on a user-defined budget, ensuring that anomalies with the highest scores are visually highlighted. To address the NP-hard nature of generating optimal plots, the authors propose an approximation algorithm. However, none of the above methods, including~\cite{gupta2018beyond} and \cite{samek2016evaluating}, perform qualitative evaluation of explanations in the absence of ground truth. Furthermore, some approaches are model-dependent, meaning the quality of generated explanations is influenced by the underlying model’s accuracy.

Our method, EiForest, leverages Isolation Forest (iForest) as a data structure while extracting additional novel features, unlike iPath~\cite{vinh2016discovering}, which relies solely on path length as the scoring mechanism for subspaces. Using only path length restricts the correctness of explanations to the accuracy of the iForest algorithm. In contrast, our EMI method generates a rule set that identifies a subspace in the $m$-dimensional space where the anomalous point is most isolated, without requiring any learning. This differs from Explainer~\cite{kopp2020anomaly}, where rules are expressed in disjunctive form and decision trees are trained on imbalanced data.

\textbf{Adverse Remarks in Audit Reports}: Identifying adverse remark sentences from a corpus of audit reports is essentially a sentence classification task. We employ FinBERT, which has been widely used in financial applications; for instance, \cite{yadav2019sentiment} utilized it for sentiment analysis on financial news to study the impact of public sentiment on stock prices. These models typically classify sentences as positive, negative, or neutral. However, negative sentiment does not always indicate an adverse remark—for example, company has incurred loss of Rs. 320 lacs reflects a financial fact, not an adverse remark, despite its negative sentiment. Our Task B resembles Extreme Multi-Label Classification (XMLC), as there are 1,437 XBRL categories, and a single sentence may correspond to multiple categories. Supervised ML approaches are challenging due to the significant time and effort required from expert accountants to create labeled training data. SurveyCoder~\cite{PP13},~\cite{patil2015active} propose unsupervised methods for assigning one or more codes from a predefined code-frame to customer responses in open-ended questions by checking semantic unit overlaps. Similarly, \cite{devine2022unsupervised} presents an unsupervised XMLC approach for tagging tech forum posts (e.g., StackOverflow) using embeddings of posts and category tags from Transformer-based models, comparing them via cosine similarity and reporting tags above a threshold. \cite{loukas2022finer} models XBRL tagging as a named entity recognition task but uses a supervised approach and considers only 139 categories, whereas we address all 1,437 categories to develop an interpretable, unsupervised method.

\section{Datasets}\label{sec:datasets}

\begin{center}
\small
\begin{longtable}{|p{0.07\textwidth}|p{0.17\textwidth}|p{0.27\textwidth}|p{0.49\textwidth}|}
\hline
\textbf{Name} & \textbf{Size} & \textbf{+1\%} &  \textbf{Description}
\\ \hline\hline
$D_{ann}^{(1)}$ & \#sent=300 & 42.0 ADV\_REMARK\% & Adverse remarks manually annotated 
\\\hline
$D_{ann}^{(2)}$ & \#sent=500 & 16.4 ADV\_REMARK\% & Adverse remarks manually annotated 
\\\hline
$D_{0}$ & \#sent=13698 & 20.7 ADV\_REMARK\% & Adverse remarks using linguistic rules
\\\hline
$D_{1}$ & \#sent=1152 & 47.6 ADV\_REMARK\% & Adverse remarks using MISTRAL-7B-Instruct\footnote{https://huggingface.co/mistralai/Mistral-7B-Instruct-v0.3}
\\\hline
$D_{str}$ & \#companies= 11460 & - & 2292 companies for 5 years, 42 financial variables + 22 ratios
\\\hline
$D_{str}^{(SL)}$ & \#labels=11460 & 19.8 MISINFO\% &  Silver labels for misinformation 
\\\hline
$D_{aud}$ & \#reports= 11460 & - & Audit reports of 2292 companies for 5 years
\\
\hline
\caption{Summary of datasets.}
\label{tab:dataset_summ}
\end{longtable}
\end{center}

We have web-scrapped the semi-structured Financial Statements (FS) - mainly Balance Sheets (BS), Income Statements (IS) and Cash-flow Statements (CS) - of 2292 listed Indian companies for five years (2010 to 2014)\footnote{https://www.moneycontrol.com/\label{money-control}}. We extracted a total of 42 {\em financial variables} - or just {\em variables} - (28 from BS, 9 from IS and 5 from CS) and 22 {\em financial ratios}, which we used to find companies with misinformation in their FS. Tables~\ref{tab:var_stat_2014},~\ref{tab:ratio_stat_2014} contain the description of the financial variables and ratios used, respectively. Note that the data does not contain any labels regarding whether a FS contains any kind of misinformation or not. This makes using supervised techniques inappropriate. This raw, structured dataset is denoted as $D_{str}$, where each of $2292 \times 5 = 11,460$ rows consists of 64 (42+22) features for a particular company for a particular year. 

\begin{figure}[!ht]
\footnotesize
\centering
\begin{subfigure}[b]{0.45\textwidth}
\includegraphics[width=1.1\linewidth]{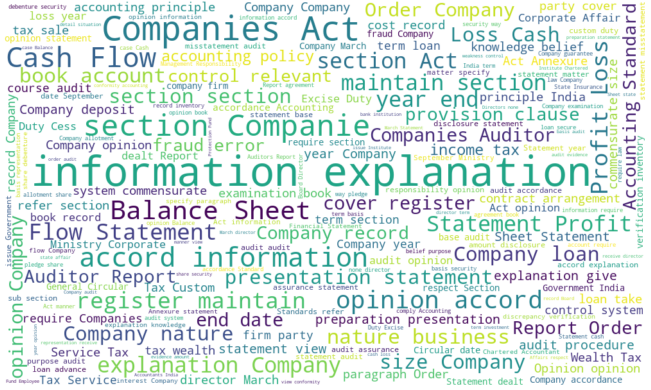}
\caption{Word cloud: Nouns}
\label{img:nouns_wc}
\end{subfigure}
\hfill
\begin{subfigure}[b]{0.45\textwidth}
\includegraphics[width=1.1\linewidth]{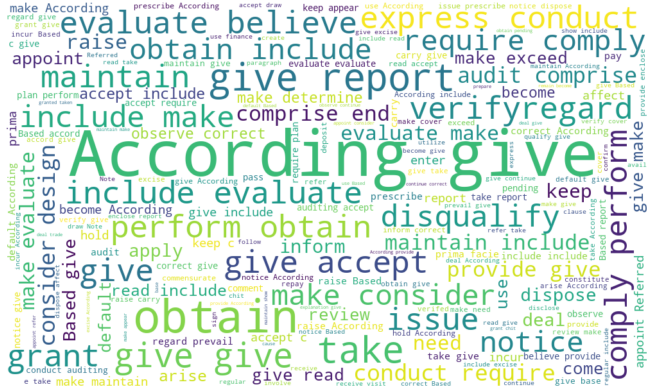}
\caption{Word cloud: Verbs}
\label{img:verbs_wc}
\end{subfigure}
\caption{Word clouds for 5\-years audit corpus.}
\end{figure}


Audit reports (as text files) are available for each of these companies for all 5 years. We have web-scrapped\textsuperscript{\ref{money-control}} the audit reports of these companies for the same 5-year time span. A company's financial health and the accuracy of information in its FS are evaluated by the auditors in their audit reports. For the 2292 companies for 5 years, there were a total of $2292 \times 5 = 11460$ audit reports. The raw corpus of audit reports is denoted $D_{aud}$. After cleaning and removing small and non-sentence lines (e.g., tables), $D_{aud}$ contains 578789 sentences and 22233701 words. The average number of sentences and words per audit report in the dataset $D_{aud}$ are 50.50 and 1891.40 respectively. The Figures \ref{img:nouns_wc} and \ref{img:verbs_wc} represent the word clouds of \textit{nouns} and \textit{verbs} from $D_{aud}$ respectively. 

\begin{figure}[h]
\footnotesize
    \centering
    \includegraphics[width = 0.7\linewidth]{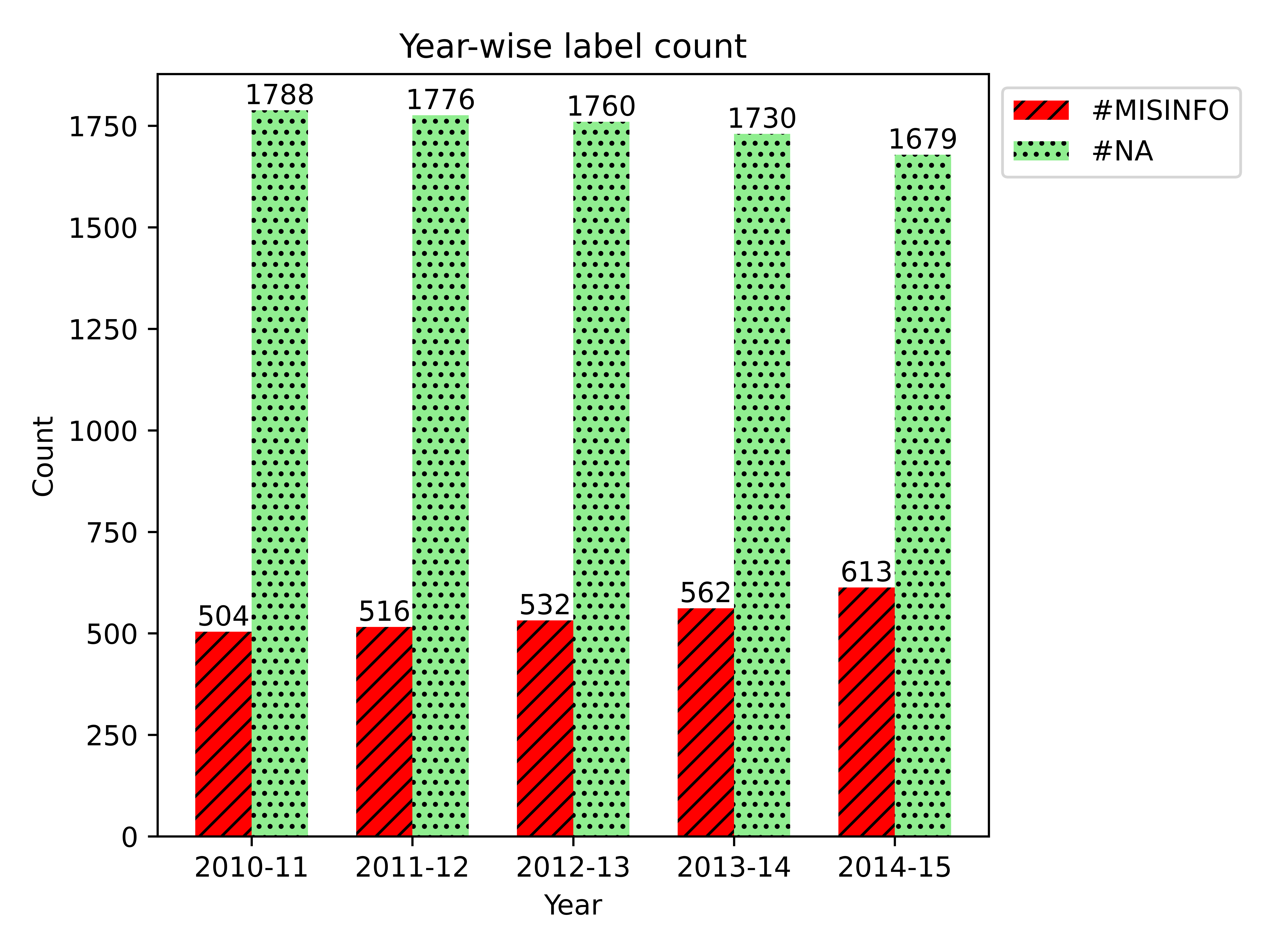}
    \caption{Year-wise label count in $D_{str}^{(SL)}$}
    \label{fig:silverLabels_count}
\end{figure}

Our focus is to extract (from each audit report) sentences that indicate {\em adverse remarks} by the auditors. These adverse remarks provide crucial help in saying whether the associated FS might contain misinformation or not. We created several datasets from these two raw datasets (Table ~\ref{tab:dataset_summ}).
\begin{itemize}
\item \textbf{Dataset $D_{0}$:} We used linguistic rules from~\cite{PAPV23} to automatically label each sentence from the dataset $D_{aud}$ as {\small{\sf ADV\_REMARK}} or {\small{\sf NA}}. This resulted 13698 sentences (2.37\%) labeled as {\small{\sf ADV\_REMARK}} and the rest as {\small{\sf NA}}. 
\item \textbf{Dataset $D_{1}$:} We used the trained adverse remark classification BERT model from ~\cite{PAPV23} to predict labels of each sentence in $D_{aud}$. This resulted in 33308 sentences (5.75\%) labeled as {\small{\sf ADV\_REMARK}} and the rest as {\small{\sf NA}}. In an attempt to select the most diverse sentences as adverse remarks, we clustered these 33308 sentences and selected 1 representative sentence from each cluster; this yielded 1152 sentences. For clustering, we used a \textit{Fast Clustering} algorithm from sentence-transformers python library\cite{reimers-2019-sentence-bert}. Then we used the MISTRAL-7B-Instruct LLM to label these 1152 sentences in the dataset $D_{aud}$ as {\small{\sf ADV\_REMARK}} or {\small{\sf NA}}. The prompt given to the LLM for classification was as follows:\\
            {\small {\tt Check whether the following sentence from a financial audit report is an adverse remark.\\
            Sentence:$S$ \\
            Answer as "Yes" or "No" followed by a short explanation.\\
            Answer:}} \\and if the output is {\sf Yes} then we label it as {\small {\sf ADV\_REMARK}}, else {\small {\sf NA}}.
\item \textbf{Dataset $D_{ann}^{(1)}$:} This dataset consists of 300 randomly selected sentences from $D_{aud}$, each labelled manually as either {\small{\sf ADV\_REMARK}} or {\small{\sf NA}}. The annotation guidelines are explained below.
\item \textbf{Dataset $D_{ann}^{(2)}$:} For obtaining unseen sentences to evaluate sentence classification models, we used audit reports from 2015, removed the sentences that did not have any verb, or had length less than 10 words or greater than 50 words and from the remaining sentences, randomly selected 500 sentences making sure they are sufficiently different from each other to avoid near-duplicates. We designed detailed annotation guidelines using which we manually labelled each sentence as either {\small{\sf ADV\_REMARK}} or {\small{\sf NA}}. A total of 82 sentences (16.4\%) were labeled as {\small{\sf ADV\_REMARK}} and the remaining 418 as {\small{\sf NA}}.
\item \textbf{Dataset $D_{str}^{(SL)}$:} We do not have gold-standard ground truth about misinformation for any company for any year. So to enable evaluation of our methods on misinformation detection, we created a silver-standard dataset, where for each company and each year, the label indicates whether its FS for that year contains misinformation or not. The method to assign such a predicted silver label is as follows. Let $A$ denote the audit report for a company $C$ for year $Y$. We use MISTRAL-7B-Instruct to label each sentence in $A$ as {\small {\sf ADV\_REMARK}} or {\small {\sf NA}}. We also use our specially-trained T5 classifier (explained below) to label each sentence in $A$ as {\small {\sf ADV\_REMARK}} or {\small {\sf NA}}. If $A$ contains at least 2 sentences both of which are labeled as {\small {\sf ADV\_REMARK}} by {\em both} MISTRAL and T5 classifier, then we assign the label {\small {\sf MISINFO}} (contains misinformation) to the tuple $(C, Y)$; else we assign it the label {\small {\sf NA}}. We denote this dataset as $D_{SL}$. Using this method, out of 11460 companies (over 5 years), $2727$ companies (19.80\%) were labeled as containing misinformation. Year-wise distribution of the silver labels is given in Figure \ref{fig:silverLabels_count}.
\end{itemize}

A brief highlight of the annotation guidelines for marking a sentence as {\small{\sf ADV\_REMARK}} are as follows: Adverse remarks are sentences which contain mentions of the detected or potential instances of non-conformance, misinformation, irregularities, inconsistencies, errors, inaccuracies, frauds, lapses, non-compliance, violations etc.; e.g., {\small {\tt Loss for the current year is understated by Rs. X and current liabilities are also understated by Rs. Y.}} Adverse remarks often also include auditors' suggestions for improvement. In addition, a sentence containing mention of an action of the management/state of the company which indicates a problem should be marked adverse. Example: {\small {\tt Company is not in a position to meet its financial obligations.}} A sentence in which the auditor is only stating facts (without negative opinion) should {\em not} be marked as adverse, though the facts may appear negative from a business perspective. Example: the sentence {\small {\tt Company has disclosed the impact of pending litigation in financial statements,}} should not be marked as {\small{\sf ADV\_REMARK}}, because the auditor is noting that the company had taken a correct action. Kappa statistic for inter-annotator agreement (2 annotators) was $\kappa$ = 0.718, which indicates substantial agreement. \\*

\begin{table*}[h]
\small
\centering
    \begin{tabular}{p{0.15\textwidth} |   p{0.85\textwidth}}
\hline
        Sentence & {\small \tt{In this situation , we are unable to comment upon the non provision of statutory liabilities for current year as well as for the earlier years. }} \\\hline
        Label & {\small {\sf ADV\_REMARK}}\\\hline
        Explanation from MISTRAL    &   {\small \tt{This sentence is an adverse remark because it indicates that the company has not made sufficient provisions for statutory liabilities, which could result in under reported expenses and potential financial instability. The auditor is unable to provide an opinion on the accuracy of the financial statements due to this significant deficiency.}} \\ \hline
        {\sf input\_text} & {\small \tt{Check whether the following sentence from a financial audit report contains an adverse remark. Answer as "adverse\_remark" or "NA" followed by a short explanation in 1 sentence. Sentence: In this situation , we are unable to comment upon the non provision of statutory liabilities for current year as well as for the earlier years. Answer:}} \\ \hline
        {\sf output\_text} & {\small \tt{Label: {\small {\sf adv\_remark}}
Explanation: This sentence is an adverse remark because it indicates that the company has not made sufficient provisions for statutory liabilities, which could result in under reported expenses and potential financial instability. The auditor is unable to provide an opinion on the accuracy of the financial statements due to this significant deficiency.}} \\\hline
    
    \end{tabular}
    \caption{Example of training data used for T5 model.}
    \label{tab:t5-train-example}
\end{table*}

\textbf{Training of T5 classifier:} We now explain how we trained the T5 classifier to label a given sentence as {\small {\sf ADV\_REMARK}} or {\small {\sf NA}}. 
We used sentence datasets $D_{0}$, $D_{1}$ and $D_{ann}^{(1)}$ to train the T5 model. The training process involved two steps:

\begin{enumerate}
    \item First, we generated a one-line explanation using MISTRAL-7B-Instruct for each sentence-label pair in the datasets. The prompt for generating the explanation for any sentence $S$ with label {\small {\sf ADV\_REMARK}} (or, {\small {\sf NA}}) in the dataset, is given as 
    {\small \tt{Explain why the following sentence from a financial audit report is an adverse remark (or, not an adverse remark, respectively). Keep the explanation concise and up to 2 sentences. \\
    Sentence:\{$S$\}. Explanation:}}
     \item Suppose that, for a sentence-label pair $(S,l)$ in the dataset, the generated explanation in $step 1$ is $exp$. We used the generative ability of T5 classifier (encode-decoder model) to generate text as output. The training data for the T5 classifier consists of an {\sf input\_text} and an {\sf output\_text}. The {\sf input\_text} contains a textual prompt  concatenated with the sentence $S$ where as the {\sf output\_text} contains the label for the respective sentence, $l$ and the explanation $exp$ generated in $step 1$ for $(S, l)$. Example of a sample used in training is given in Table \ref{tab:t5-train-example}.
 \end{enumerate} 
 
The T5 classifier trained on this data is capable of classifying a given audit sentence as {\sf ADV\_REMARK} or {\sf NA} and provide a one-line explanation for its classification. We evaluated the classifier on the $D_{ann}^{(2)}$ dataset. We extracted the predicted labels from the output of the classifier and evaluated the results with F1-score as evaluation metric (see Table \ref{tab:Eval_results_adv_remark_clf}). The T5 classifier has F1-score of $0.80$.

\begin{table}[]
    \small
    \centering
    \begin{tabular}{|p{0.18\textwidth}|p{0.1\textwidth}|p{0.1\textwidth}|p{0.1\textwidth}|}
    \hline
        \textbf{Model}  &   \textbf{P}  &   \textbf{R}  &   \textbf{F1} \\\hline\hline
        $Attn$-BERT\cite{PAPV23}   &   0.76    &    0.70   &   0.72    \\\hline
        T5  &    \textbf{0.85}   &   \textbf{0.76}    &   \textbf{0.80}    \\\hline
    \end{tabular}
    \caption{Evaluation results for adverse remark classification on $D_{ann}^{(2)}$}
    \label{tab:Eval_results_adv_remark_clf}
\end{table}

\normalsize
\begin{center}
\small
\begin{longtable}{|p{0.45\textwidth}|p{0.08\textwidth}|p{0.08\textwidth}|p{0.08\textwidth}|p{0.08\textwidth}|p{0.08\textwidth}|}
     \hline 
     \textbf{Variable} & \textbf{Mean} & \textbf{Stdev} & \textbf{Q1} & \textbf{Q2} & \textbf{Q3} \\\hline\hline
     $V_{1}$ Equity Share Capital &  65.76 & 407.39 & 5.10 & 10.97 & 25.25\\\hline
     $V_{2}$ Reserves and Surplus & 912.39 & 6437.14 & 1.43 & 29.12 & 213.41\\\hline
     $V_{3}$ Total Shareholders Funds & 991.00 & 6657.67 & 9.30 & 45.44 & 258.93 \\\hline
     $V_{4}$ Long Term Borrowings & 568.60 & 4930.10 & 0.00 & 3.22 & 46.05 \\\hline
     $V_{5}$ Deferred Tax Liabilities [Net] & 42.74 & 475.25 & 0.00 & 0.03 & 4.31 \\\hline
     $V_{6}$ Other Long Term Liabilities & 44.02 & 416.29 & 0.00 & 0.00 & 1.44 \\\hline
     $V_{7}$ Long Term Provisions & 31.07 & 576.92 & 0.00 & 0.10 & 1.64 \\\hline
     $V_{8}$ Total Non-Current Liabilities & 686.44 & 5334.67 & 0.50 & 9.30 & 77.31 \\\hline
     $V_{9}$ Short Term Borrowings & 221.53	& 1434.74 & 0.00 & 7.93 & 60.57\\\hline
     $V_{10}$ Trade Payables & 209.32	& 1667.88	& 0.32 & 8.99& 54.12\\\hline
     $V_{11}$ Other Current Liabilities & 245.01	& 1283.46 &	0.44	& 6.41	& 50.16\\\hline
     $V_{12}$ Short Term Provisions & 59.62	& 642.18 &	0.06 &	1.00 &	7.70 \\\hline
     $V_{13}$ Total Current Liabilities & 735.47 & 4337.37	& 4.00	& 40.50	& 231.55\\\hline
     $V_{14}$ Total Capital And Liabilities &  2417.84	& 14339.12	& 22.84	& 124.10	& 611.38\\\hline
     $V_{15}$ Fixed Assets &  787.58 & 6084.09 & 2.09 & 25.05 & 154.91\\\hline
     $V_{16}$ Non-Current Investments & 304.64 & 2191.23 & 0.00 & 1.89 & 28.12\\\hline
     $V_{17}$ Deferred Tax Assets [Net] &  4.92 & 57.39 & 0.00 & 0.00 & 0.00\\\hline
     $V_{18}$ Long Term Loans And Advances &  392.88 & 5210.24 & 0.22 & 2.76 & 20.69\\\hline
     $V_{19}$ Other Non-Current Assets &  26.63 & 305.05 & 0.00 & 0.00 & 0.60\\\hline
     $V_{20}$ Total Non-Current Assets &  1516.65 & 10130.76 & 9.02 & 49.88 & 303.29\\\hline
     $V_{21}$ Current Investments &  80.64 & 977.09 & 0.00 & 0.00 & 0.00\\\hline
     $V_{22}$ Inventories &  224.10 & 1842.32 & 0.14 & 10.06 & 66.23 \\\hline
     $V_{23}$ Trade Receivables &  195.23	&	1023.19	&	0.85	&	14.12	&	72.03\\\hline
     $V_{24}$ Cash And Cash Equivalents & 136.10	&	1244.63	&	0.31	&	2.38	&	14.24\\\hline
     $V_{25}$ Short Term Loans And Advances &  182.56	&	1241.64	&	0.76	&	6.22	&	36.33\\\hline
     $V_{26}$ OtherCurrentAssets &  82.51	&	796.76	&	0.00	&	0.23	&	3.75\\\hline
     $V_{27}$ Total Current Assets & 901.14	&	5195.71	&	9.32	&	56.93	&	285.37\\\hline
     $V_{28}$ Total Assets &  2417.84	&	14339.12	&	22.84	&	124.10	&	611.38\\\hline
     $V_{29}$ Other Income &  46.51	&	449.75	&	0.13	&	1.06	&	6.42\\\hline
     $V_{30}$ Total Revenue &  1746.50	&	15693.73	&	7.16	&	80.63	&	469.23\\\hline
     $V_{31}$ Cost Of Materials Consumed &  726.48	&	9090.48	&	0.00	&	4.93	&	142.91\\\hline
     $V_{32}$ Changes In Inventories Of FGWIP And Stock-In Trade &  -5.05	&	87.19	&	-1.45	&	0.00	&	0.11\\\hline
     $V_{33}$ Employee Benefit Expenses &  105.82	&	895.59	&	0.48	&	5.34	&	28.72\\\hline
     $V_{34}$ Depreciation And Amortisation Expenses &  52.98	&	403.43	&	0.16	&	2.01	&	12.84\\\hline
     $V_{35}$ Other Expenses &  184.24	&	1069.26	&	0.86	&	9.06	&	58.22\\\hline
     $V_{36}$ Profit/Loss Before Tax &  152.70	&	1336.24	&	-0.10	&	1.09	&	19.14\\\hline
     $V_{37}$ Current Tax &  39.39	&	294.63	&	0.00	&	0.17	&	4.22\\\hline
     $V_{38}$ Net Profit/Loss Before Extraordinary Items And Tax &  156.39	&	1358.32	&	-0.07	&	1.06	&	19.72\\\hline
     $V_{39}$ Net CashFlow From Operating Activities & 142.94	&	1690.20	&	-0.26	&	2.28	&	28.41\\\hline
     $V_{40}$ Net Cash Used In Investing Activities &  -131.03	&	1674.14	&	-14.42	&	-0.69	&	0.10\\\hline
     $V_{41}$ Net Cash Used From Financing Activities &  -16.20	&	909.41	&	-12.87	&	-0.30	&	0.59\\\hline
     $V_{42}$ Net Inc/Dec In Cash And Cash Equivalents & -2.37	&	449.75	&	-0.95	&	0.00	&	0.64\\\hline

\caption{Financial Variables and summary statistics - 2014}
    \label{tab:var_stat_2014}
\end{longtable}
\end{center}
\normalsize

\normalsize

\section{Unsupervised Methods for Misinformation Detection}\label{sec:misinfo}

As mentioned earlier, the dataset does not have any {\em gold standard} labels for any of the FS, as to whether the FS contain any known and confirmed instances of misinformation or not. We use the adverse remarks found in the audit report associated with a company's FS to automatically build such labels; this is the $D^{SL}_{str}$ dataset (Figure ~\ref{fig:silverLabels_count}). Clearly, these labels would be at best {\em silver standard} and a positive label does not necessarily mean that there is any actual instance of misinformation. Based on our study of the dataset, we treat a FS as {\em qualified} (i.e. containing misinformation) if its audit report has at least 2 sentences that are marked a {\sf ADV\_REMARK} by both the T5 classifier and MISTRAL-7B-Instruct LLM from the audit report for either in the current year or in the next year. The reason for using current and next year audit reports is that the effects of misinformation sometimes accumulate and amplify over time, which means it may be easier for an auditor to catch it next year. 

We use several types of unsupervised algorithms on the structured dataset to detect FS in a particular year that are likely to contain misinformation. We then use the silver labels in $D^{SL}_{str}$ dataset to compute the precision of these detection methods as follows. Suppose FS of a company $C$ for year $Y$ are detected as likely containing misinformation. If the corresponding label is {\small {\sf MISINFO}} either for that year or the next year (if available), then we mark $C$ as {\em true positive} else as {\em false positive}.

\begin{center}
\begin{table*}[h]
\small
    \begin{tabular}{|l|l|p{0.1\textwidth}|p{0.1\textwidth}|p{0.1\textwidth}|p{0.1\textwidth}|p{0.1\textwidth}|}
             \hline 
     \textbf{Name} & \textbf{Formula} & \textbf{Mean} & \textbf{Stdev} & \textbf{Q1} & \textbf{Q2} & \textbf{Q3} \\\hline\hline
     $R_{1}$    &   $V_{23}$/$V_{28}$   &   0.15	&	0.16	&	0.02	&	0.11	&	0.24 \\\hline
     $R_{2}$    &   $V_{27}$/$V_{28}$   &   0.50	&	0.27	&	0.30	&	0.51	&	0.71  \\\hline
     $R_{3}$    &   $V_{15}$/$V_{28}$   &   0.28	&	0.24	&	0.05	&	0.24	&	0.44 \\\hline
     $R_{4}$    &   $\log(V_{28})$   &   4.83	&	2.44	&	3.14	&	4.83	&	6.42 \\\hline
     $R_{5}$    &   $V_{24}$/$V_{28}$   &   0.06	&	0.12	&	0.01	&	0.02	&	0.05 \\\hline
     $R_{6}$    &   $\log(V_{8}+V_{13})$    &   3.80	&	3.00	&	2.03	&	4.06	&	5.85 \\\hline
     $R_{7}$    &   ($V_{27}$-$V_{22}$)/$V_{13}$ &  10.20	&	86.77	&	0.56	&	0.97	&	2.25\\\hline
     $R_{8}$    &   ($V_{8}$+$V_{13}$)/$V_{28}$ & 0.93	&	6.47	&	0.23	&	0.51	&	0.71\\\hline
     $R_{9}$    &   $V_{13}$/$V_{27}$ &  4.03	&	50.50	&	0.35	&	0.73	&	1.01\\\hline
     $R_{10}$   &   ($V_{8}$+$V_{13}$)/$V_{3}$  &   0.86	&	38.34	&	0.16	&	0.81	&	1.95\\\hline
     $R_{11}$   &   $V_{13}$/$V_{28}$   &   0.64	&	5.63	&	0.13	&	0.32	&	0.50\\\hline
     $R_{12}$   &   $V_{24}$/$V_{27}$   &   0.13	&	0.20	&	0.02	&	0.05	&	0.13\\\hline
     $R_{13}$   &   $V_{22}$/$V_{28}$   &   0.14	&	0.16	&	0.00	&	0.10	&	0.22\\\hline
     $R_{14}$   &   $V_{36}$/$V_{30}$   &   -5.65	&	243.44	&	-0.02	&	0.03	&	0.11\\\hline
     $R_{15}$   &   ($V_{36}$-$V_{37}$)/$V_{3}$ &   0.11	&	3.26	&	0.00	&	0.04	&	0.12\\\hline
     $R_{16}$   &   ($V_{36}$-$V_{37}$)/$V_{28}$    &   -0.01	&	0.56	&	-0.01	&	0.01	&	0.05\\\hline
     $R_{17}$   &   ($V_{36}$-$V_{37}$)/$V_{30}$    &   -5.67	&	243.44	&	-0.02	&	0.02	&	0.09\\\hline
     $R_{18}$   &   $V_{39}$/$V_{13}$   &   0.24	&	14.22	&	-0.04	&	0.11	&	0.34\\\hline
     $R_{19}$   &   $V_{39}$/($V_{36}$-$V_{37}$)    &   5.01	&	204.69	&	-0.41	&	0.80	&	2.50\\\hline
     $R_{20}$   &   ($V_{7}$+$V_{12}$)/($V_{8}$+$V_{13}$)   &   0.10	&	0.19	&	0.01	&	0.03	&	0.10\\\hline
     $R_{21}$   &   ($V_{4}$+$V_{9}$)/($V_{8}$+$V_{13}$)    &   0.42	&	0.30	&	0.10	&	0.44	&	0.65\\\hline
     $R_{22}$   &   ($V_{5}$+$V_{10}$)/($V_{8}$+$V_{13}$)   &   0.26	&	0.24	&	0.06	&	0.20	&	0.38\\\hline
    \end{tabular}
    \caption{Financial ratios and summary statistics - 2014}
    \label{tab:ratio_stat_2014}
\end{table*}
    
\end{center}

\subsection{Anomaly Detection (AD)}

One reasonable hypothesis is that any FS which contains some misinformation would appear anomalous in some well-defined sense, as compared to {\em honest} FS (i.e., one without any misinformation). To test this idea, we used several well-known {\em Anomaly Detection} (AD) algorithms implemented in PyOD package~\cite{ZLN19} on the structured dataset to identify anomalous (i.e., suspicious) FS. The ensemble method takes top-20 FS having the highest count of AD algorithms which marked it as anomalous (we do this year-by-year). 

Table~\ref{tab:anom_det_full} shows the precision@20 (P@20) results for 18 (17 + 1 ensemble) AD algorithms. As seen, Connectivity-Based Outlier Factor (COF) algorithm~\cite{TCWC01} has the highest P@20 (0.60) for year 2014 i.e., 12 out of 20 companies identified by it as anomalous are indeed marked as {\small {\sf MISINFO}} in the $D_{str}^{(SL)}$. The P@20 results of all the algorithms for years 2010-2014 can be seen in the Table \ref{tab:anom_det_full}. The $averageP@20$ column shows the average precision of an algorithm across 5 years. Notice that the Linear Outlier Factor (LOF)\cite{zhong2019steganographer}  algorithm, gives the highest precision for 4 (out of 5) years in the data. The LOF algorithm computes the local density of each point based on the distance between its k-nearest neighbors. The points with less density are given as outliers. 
Other algorithms such as COF and SOS have the high precision for the year 2014 and give a descent average precision across all years respectively. On the other hand, the Histogram Based Outlier Detection (HBOS)\cite{goldstein2012histogram} algorithm preforms poorly for all the years. HBOS constructs histograms for each feature with fixed or dynamic number of bins (depending on data distribution) and decides the outliers by estimating the density based on the frequency of samples in each bin. Lower the density indicates higher the likelihood of being an outlier.

\subsection{Step-wise Regression}\label{subsec:regr}

When a FS for a company is identified as suspicious (i.e., likely to contain misinformation), it is important to provide a justification or explanation for this prediction, which is human-understandable. For improved usability, the key question to be answered is: \textit{where exactly is the misinformation present in a FS?} It is well-known that some financial variables in the FS - e.g., those related to assets and liabilities - are more susceptible for misinformation than others; e.g., auditors adverse remarks are more often based on these variables\cite{SVAP22}. For example, misinformation often tends to deflate losses or liabilities or inflate profits. 

\begin{table}[ht]
\small
\centering
\begin{tabular}{|l|l|l|l|l|l|l|}
\hline
\textbf{Algorithm} & \textbf{2010} & \textbf{2011} & \textbf{2012} & \textbf{2013} & \textbf{2014}  &   \textbf{averageP@20}\\ \hline\hline
ABOD	&	0.30	&	0.25	&	0.20	&	0.25	&	0.35    &   0.27    \\	\hline
CBLOF	&	0.30	&	0.25	&	0.15	&	0.25	&	0.25    &   0.24\\	\hline
COF	&	0.40	&	0.60	&	0.30	&	0.35	&	\textbf{0.60}    &   0.45   \\	\hline
HBOS	&	0.10	&	0.25	&	0.10	&	0.15	&	0.20    &   0.16\\	\hline
iForest	&	0.15	&	0.20	&	0.15	&	0.20	&	0.25    &   0.19\\	\hline
KNN	&	0.30	&	0.25	&	0.15	&	0.20	&	0.20    &   0.22\\	\hline
LMDD	&	0.20	&	0.15	&	0.30&	0.35	&	0.40    &   0.28\\	\hline
LODA	&	0.25	&	0.20	&	0.10&	0.30	&	0.20    &   0.21\\	\hline
LOF	&	\textbf{0.65}	&	\textbf{0.65}	&	\textbf{0.55}	&	\textbf{0.40}	&	0.55    &   \textbf{0.56}\\	\hline
Mahalanobis	&	0.35	&	0.30	&	0.45	&	0.40	&	0.30    &   0.36\\	\hline
MCD	&	0.30	&	0.25	&	0.20	&	0.25	&	0.40    &   0.28\\	\hline
OCSVM	&	0.20	&	0.30	&	0.25	&	0.00	&	0.15    &   0.18\\	\hline
PCA	&	0.35	&	0.25	&	0.30	&	0.30	&	0.30    &   0.30\\	\hline
SOD	&	0.30	&	0.30	&	0.30	&	0.25	&	0.25    &   0.28\\	\hline
SOS	&	0.40	&	0.50	&	0.50	&	0.25	&	0.45    &   0.42\\	\hline
Auto-Encoder		&	0.35	&	0.25	&	0.30	&	0.30	&	0.30    &   0.30\\	\hline
VAE	&	0.35	&	0.25	&	0.30	&	0.30	&	0.30    &   0.30\\	\hline\hline
Ensemble	&	0.35    &   0.25	&   0.25	&   0.35	&   0.30    &   0.30\\	\hline	
\end{tabular}
\caption{Anomaly detection methods - Results P@20}
\label{tab:anom_det_full}
\end{table}
\normalsize

We used the first approach from Model-based anomaly detection method from \cite{SVAP22}, where the models are various step-wise regression models. To illustrate the approach, we have considered the variables related to liabilities as {\em susceptible} variables (we could easily add other kinds of variables also). We build regression models and use them to identify suspicious FS. We start with a given susceptible variable as the {\em dependent} variable, identify other variables on which that susceptible variable depends on ({\em independent variables}), and use only these variables to build regression model(s) for that suspicious variable. Finally, we use these regression models to identify suspicious FS. The approach consists of the following steps: 

\begin{enumerate}
    \item Select a susceptible variable (say, $Y$);
    \item Identify variables most correlated with $Y$ as dependent variables; 
    \item Use step-wise regression to incrementally build multiple {\em Ordinary Least Squares} (OLS) regression models for $Y$ and select the best regression model $M_{Y}$ for $Y$ having the highest adjusted $R^{2}$ value;
    \item Use $M_{Y}$ to identify suspicious BS as follows. Take a BS, predict the value of $Y$ using the values (in this BS) of the independent variables used in $M_{Y}$, compute the prediction error $Y - \hat{Y}$ and mark those 20 BS as suspicious which have the highest squared prediction error in $Y$ using $M_{Y}$.
\end{enumerate}

\begin{description}
\item[$M14$:] $R_{8}$ $\leftarrow$ [$R_{12}$ $R_{5}$ $R_{3}$ $V_{40}$ $V_{41}$ $V_{32}$ $R_{15}$ $V_{42}$ $R_{19}$ $V_{18}$ $V_{17}$ $V_{19}$ $V_{31}$ $V_{26}$ $V_{22}$ $V_{30}$ $R_{1}$ $R_{14}$ $R_{17}$ $V_{1}$ $V_{15}$ $V_{21}$ $V_{25}$ $V_{39}$ $V_{20}$ $V_{24}$ $V_{29}$ $V_{34}$ $V_{27}$ $V_{28}$ $V_{33}$ $V_{16}$ $V_{23}$ $V_{35}$ $V_{36}$ $V_{37}$ $V_{38}$ $V_{2}$ $V_{3}$ $R_{2}$ $R_{13}$ $R_{4}$ $R_{16}$] 
\item[$M15$:]$R_{9}$ $\leftarrow$  [$R_{12}$ $V_{1}$ $R_{3}$ $V_{40}$ $V_{32}$ $V_{41}$ $V_{42}$ $R_{15}$ $R_{19}$ $V_{18}$ $V_{19}$ $R_{14}$ $R_{17}$ $V_{17}$ $V_{31}$ $V_{39}$ $V_{21}$ $V_{26}$ $V_{15}$ $V_{16}$ $V_{20}$ $V_{24}$ $V_{29}$ $V_{30}$ $V_{34}$ $V_{22}$ $V_{33}$ $V_{36}$ $V_{38}$ $V_{3}$ $V_{25}$ $V_{28}$ $V_{37}$ $V_{2}$ $V_{35}$ $V_{27}$ $V_{23}$ $R_{5}$ $R_{4}$ $R_{1}$ $R_{13}$ $R_{16}$ $R_{2}$] 
\item[$M16$:]$R_{10}$    $\leftarrow$   [$R_{4}$ $R_{3}$ $R_{1}$ $V_{18}$ $V_{25}$ $V_{28}$ $V_{20}$ $V_{23}$ $V_{27}$ $R_{5}$ $R_{12}$ $R_{16}$ $V_{15}$ $V_{22}$ $V_{26}$ $V_{30}$ $V_{34}$ $V_{41}$ $V_{31}$ $V_{1}$ $V_{19}$ $V_{35}$ $R_{14}$ $R_{17}$ $V_{2}$ $V_{3}$ $V_{16}$ $V_{24}$ $V_{37}$ $V_{21}$ $V_{29}$ $V_{33}$ $V_{36}$ $V_{38}$ $V_{42}$ $V_{17}$ $V_{32}$ $V_{40}$ $V_{39}$ $R_{19}$ $R_{2}$ $R_{13}$ $R_{15}$]
\item[$M17$:]$R_{11}$    $\leftarrow$  [$R_{12}$ $R_{1}$ $R_{5}$ $V_{40}$ $V_{32}$ $V_{42}$ $R_{3}$ $R_{15}$ $V_{41}$ $R_{19}$ $V_{31}$ $V_{17}$ $V_{19}$ $V_{22}$ $V_{26}$ $V_{30}$ $V_{39}$ $V_{18}$ $V_{21}$ $V_{24}$ $V_{25}$ $V_{29}$ $V_{33}$ $R_{14}$ $R_{17}$ $V_{1}$ $V_{15}$ $V_{23}$ $V_{34}$ $V_{16}$ $V_{27}$ $V_{35}$ $V_{36}$ $V_{37}$ $V_{38}$ $V_{20}$ $V_{28}$ $V_{2}$ $V_{3}$ $R_{2}$ $R_{13}$ $R_{4}$ $R_{16}$]
\end{description}
\captionof{table}{Best Regression Models - 2014}
\label{tab:best_regression_models_2014}

Apart from the OLS regression, we used the models with same input-output structure but built on Lasso, Ridge, Support Vector Regression (SVR) and Random Forest Regression (RFR). Lasso (or Ridge) is a regularization technique that supplements the OLS formulation by adding $L1\ penalty$ (or $L2\ penalty$, respectively) to the objective function. SVR (based on Support Vector machines algorithm) is used to model the complex, non-linear relationships in the data and predict the continuous outputs for the given inputs. RFR fits a number of decision tree regressors on various sub-samples of the dataset and uses averaging to improve the predictive accuracy and control over-fitting \cite{regressionRFR}. To illustrate, Table~\ref{tab:best_regression_models_2014} shows the 4 best regression models built using step-wise OLS regression for susceptible variables related to liabilities: $R_8$, $R_9$, $R_{10}$, $R_{11}$. We used a total of $21$ variables and ratios that are related to liabilities, as susceptible variables and built regression models. The P@20 results for these regression models on the 5-year data are given in Table~\ref{tab:step_reg}. As we can see, the SVR models have higher precision as compared to the other models. 

\begin{table*}[h]
    \centering
    \small
    \begin{tabular}{|p{0.13\textwidth}|p{0.11\textwidth}|p{0.11\textwidth}|p{0.11\textwidth}|p{0.11\textwidth}|p{0.11\textwidth}|p{0.12\textwidth}|}
        \hline
\textbf{Model}	&	 \textbf{2010-P@20}	&	 \textbf{2011-P@20}	&	 \textbf{2012-P@20}	&	 \textbf{2013-P@20}	&	 \textbf{2014-P@20}	&   \textbf{average-P@20}   \\\hline\hline
OLS M14 	&	0.65	&	 \textbf{0.70}	&	 \textbf{0.70}	&	 \textbf{0.65}	&	 \textbf{0.70}  &   \textbf{0.68}\\
OLS M15 	&	0.55	&	0.55	&	0.5	&	0.6	&	0.45    &   0.53\\
OLS M16 	&	0.65	&	0.65	&	0.65	&	0.50	&	0.50    &   0.59\\
OLS M17 	&	\textbf{0.70}	&	0.60	&	0.65	&	0.60	&	 \textbf{0.70}	 &  0.65\\ \hline
Ridge M14 	&	0.65	&	 \textbf{0.75}	&	 \textbf{0.70}	&	 \textbf{0.65}	&	 \textbf{0.70}	&   \textbf{0.69}\\
Ridge M15 	&	0.55	&	0.60	&	0.50	&	0.60	&	0.45	&   0.54\\
Ridge M16 	&	0.65	&	0.65	&	0.65	&	0.50	&	0.50	&	 0.59\\
Ridge M17 	&	\textbf{0.70}	&	0.60	&	0.65	&	0.60	&	 \textbf{0.70}  &	 0.65\\\hline
Lasso M14 	&	0.65	&	0.70	&	 \textbf{0.80}	&	 \textbf{0.70}	&	 \textbf{0.70}	&   \textbf{0.71}\\
Lasso M15 	&	0.55	&	0.65	&	0.50	&	 \textbf{0.70}	&	0.50	&   0.58\\
Lasso M16 	&	0.65	&	0.65	&	0.65	&	0.50	&	0.50	&   0.59\\
Lasso M17 	&	\textbf{0.70}	&	 \textbf{0.75}	&	0.65	&	0.60	&	 \textbf{0.70}	&   0.68\\\hline
SVR M14 	&	\textbf{0.80}	&	0.70	&	 \textbf{0.90}	&	0.75	&	0.75	&   0.78\\
\textbf{SVR M15}	&	0.55	&	0.60	&	0.70	&	 \textbf{0.75}	&	 \textbf{0.95} 	&   0.71\\
SVR M16 	&	0.65	&	0.65	&	0.80	&	0.55	&	0.55	&   0.64\\
SVR M17 	&	\textbf{0.80}	&	 \textbf{0.90}	&	 \textbf{0.90}	&	0.75	&	0.80	&   \textbf{0.83}\\\hline
RFR M14 	&	0.50	&	 \textbf{0.50}	&	0.40	&	0.45	&	0.50	&   0.47\\
RFR M15 	&	\textbf{0.65}	&	0.35	&	0.55	&	 \textbf{0.60}	&	 \textbf{0.60}	&   0.55\\
RFR M16 	&	0.55	&	 \textbf{0.50}	&	 \textbf{0.60}	&	0.50	&	0.50	 &  \textbf{0.53}\\
RFR M17 	&	\textbf{0.65}	&	 \textbf{0.50}	&	0.45	&	0.40	&	0.55	 &  0.51\\ \hline
    \end{tabular}
    \caption{Results of Regression Models in Identifying Suspicious FS.}
\label{tab:step_reg}
\end{table*}

\section{Explanations for Misinformation}\label{sec:expl}

For further processing of detected instances of misinformation in a financial statement, it is often useful to automatically generate human-understandable justifications or explanations for each instance. Such explanations assist in determining whether the company’s financial statement truly contains misinformation or if it is simply a false alarm. Research in Explainable AI (XAI)~\cite{TSP22} focuses on producing explanations for predictions made by AI models, primarily classification models.

\begin{table*}[ht]
    \centering
    \small
    \begin{tabular}{l   l}
    \hline
\textbf{Name} & \textbf{Description} \\ \hline\hline
$f_1$ & Average depth of the trees \\
$f_2$ & Average size of the leaf containing $z$ \\
$f_3$ & Number of paths that contain feature $v$ \\
$f_4$ & Average \% drop in the partition after split \\
$f_5$ & Number of short paths (less than the maximum tree depth) \\
$f_6$ & The level at which feature $v$ is present on average \\
$f_7$ & Average \% drop in the partition after split for short paths \\
$f_8$ & The level at which feature $v$ is present in short paths on average \\
\hline
    \end{tabular}
    \caption{Summary feature vector for EiForest} \label{tab:iforest_summary}
\end{table*}

\begin{algorithm}
\small
\DontPrintSemicolon
\SetKwInOut{Input}{input}\SetKwInOut{Output}{output}
\Input{$D, V, z, k_0, c; $ s.t. $ 1 \le k_0 \le |V|; c = 1.0$}
\Output{$E_{set}$ s.t. for each $ \phi \in E_{set}$, $\phi \subseteq V$}
\Begin{
	$E_{set} = \emptyset$\;
	\For{$k=k_0$ \KwTo $0$}{
		\ForEach{$\phi \in 2^V$ and $|\phi| = k$}{ 
			\ForEach{$z \in D$}{
				$d_z = R_V(z) - R_{V \backslash \phi}(z)$;
			}
			\If{$d_z > 0$ and $d_z > \mu + c \cdot \sigma$}{
				$E_{set} = E_{set} \cup \phi$;
			}
		}
	}
	\Return $E_{set}$
}
\caption{EMD}\label{IR}
\end{algorithm}

XAI aims to provide explanations in various forms, such as weighted or non-weighted subsets of features, rule sets, visual representations, and natural language descriptions. In contrast, the research area of Outlying Aspect Mining (OAM)~\cite{SM22},~\cite{LBS23} focuses on generating explanations for the anomalousness of a point, typically as a subset of features forming a subspace. While XAI explains the reasoning of an underlying model OAM offers a holistic, model-agnostic view of why a point is interesting or anomalous.

\small
\begin{longtable}[ht]{p{1\columnwidth}}

\hline
\textbf{Parameters:}\\
\hline
$\bullet ~m$: Number of features in the dataset \\
$\bullet ~n$: Number of points in the dataset \\
$\bullet ~\mathbf{z}$: The anomalous point to be explained \\
$\bullet ~L$: Maximum number of features to be included in the explanation \\ 
$\bullet ~M_1$: $n\times m$ size matrix representing whether other points have higher values than the anomalous point\\
	$\hspace{4mm}\bullet ~M_1[j,i]=1$ only if $i^{th}$ feature of $j^{th}$ point is greater than $\mathbf{z}[i]$\\
	$\hspace{4mm}\bullet ~M_1[j,i]=0$ otherwise\\
$\bullet ~M_2$: $n\times m$ size matrix representing whether other points have lower values than the anomalous point\\
	$\hspace{4mm}\bullet ~M_2[j,i]=1$ only if $i^{th}$ feature of $j^{th}$ point is less than $\mathbf{z}[i]$\\
	$\hspace{4mm}\bullet ~M_2[j,i]=0$ otherwise\\
\hline
\textbf{Variables:}\\
\hline
$\bullet ~\mathbf{x_1}$: $m$ length binary array such that $\mathbf{x_1}[i] = 1$ implies that the $i^{th}$ feature is included in the explanation as $v_i\leq z[i]$\\
$\bullet ~\mathbf{x_2}$: $m$ length binary array such that $\mathbf{x_2}[i] = 1$ implies that the $i^{th}$ feature is included in the explanation as $v_i\geq z[i]$\\
$\bullet ~\mathbf{y}$: $n$ length array such that:\\
	$\hspace{4mm}\bullet ~\mathbf{y}[j] = 1$ only if $\exists_i ((M_1[j,i] = 1) \land (\mathbf{x_1}[i] = 1)) \lor ((M_2[j,i] = 1) \land (\mathbf{x_2}[i] = 1))$ \newline
	({\it\small $\mathbf{y}[j]$ is 1 only if $j^{th}$ point breaks at least one condition used in the explanation})\\
	$\hspace{4mm}\bullet ~\mathbf{y}[j] = 0$ otherwise ({\it\small $\mathbf{y}$ need not be an integer})\\
	
\hline
\textbf{Objective:}\\
\hline
$\hspace{5mm} \bullet ~Maximize \sum_j\mathbf{y}[j]$ ({\it\small maximize no. of other points which do not satisfy at least one condition used in the explanation})\\
\hline
\textbf{Constraints:}\\
\hline
$\bullet ~C_1$: $\sum_{i=1}^m (\mathbf{x_1}[i] + \mathbf{x_2}[i]) \leq L$ ({\small\it At most $L$ variables can be chosen in final explanation})\\

$\bullet ~C_2$: $\mathbf{x_1}[i] + \mathbf{x_2}[i] \leq 1$,  $\forall_i$ s.t. $1\leq i\leq m$ \newline
({\small\it A variable should not be repeated in the set of L variables used for the explanation.}) \\

$\bullet ~C_3$: $\mathbf{y}[j] \geq \mathbf{x_1}[i] \cdot M_1[j, i]$, $\forall_{ij}$ s.t. $1\leq i\leq m, 1\leq j\leq n$ \newline
({\small\it $\mathbf{y}[j]$ has to be at least 1 if $M_1[j, i]$ is 1 for any feature $i$ which is included in the explanation.})\\

$\bullet ~C_4$: $\mathbf{y}[j] \geq \mathbf{x_2}[i] \cdot M_2[j, i]$, $\forall_{ij}$ s.t. $1\leq i\leq m, 1\leq j\leq n$ \newline
({\small \it $\mathbf{y}[j]$ has to be at least 1 if $M_2[j, i]$ is 1 for any feature $i$ which is included in the explanation.})\\

$\bullet ~C_5$: $\mathbf{y}[j] \leq \sum_{i=1}^m (\mathbf{x_1}[i] \cdot M_1[j, i] + \mathbf{x_2}[i] \cdot M_2[j, i])$, $\forall_j$ s.t. $1\leq j\leq n$ \newline
({\small \it $\mathbf{y}[j]$ should remain 0 for the points which do not contain 1 for any of the selected variables in $M_1[j]$ and $M_2[j]$.})\\

$\bullet ~C_6$: $\mathbf{y}[j] \leq 1$, $\forall_j$ s.t. $1\leq j\leq n$ ({\small \it $\mathbf{y}[j]$ should be at most 1.})\\
\hline

\caption{ILP formulation for generating explanations}
\label{tab:ILP}
\end{longtable}
\normalsize

\subsection{Problem Formulation}

We have an $m$-dimensional dataset $D = \{x_1, x_2, \ldots, x_n\}$ where each $x_i \in \mathbf{R}^m$ and $V = \{V_1, V_2, \ldots, V_m\}$ is the {\em feature set} (names of features). Suppose we are given an anomalous instance $z \in D$,  which is obtained by some AD technique unknown to us. The objective is to generate an explanation $E$ that explains why the the data-point should be considered anomalous. While explanations can take many forms, typically $E$ is either a subset of features $E \subseteq V$ or it is a set of logical rules over $V$. In our case, $D$ is a dataset of FS of $n$ companies, where each company is represented as an $m$-dimensional feature vector and $z$ is an anomalous company that is suspected of having misinformation in its FS.

We discuss 3 methods for generating explanation for why a point could be anomalous; the methods are the same as in ~\cite{VSPP22}. However, they had used manually constructed explanations for evaluations of automatically generated explanations. In this paper, we propose a novel technique for automatically creating silver standard explanations. Moreover, we use a much larger dataset for evaluation of the algorithms. 

\subsection{Explanation using Mahalanobis Distance (EMD)}

We sort all points in $D$ in descending order of their Mahalanobis distance from the mean vector of $D$. The Mahalanobis rank $R_V(x)$ of a point $x \in D$ is defined as its position in this sorted list. For any proper subset $A \subset V$ of features, $R_{V \setminus A}(x)$ is similarly defined, except that Mahalanobis distances are computed after removing all features in $A$ from every point in $D$. Note that a smaller rank value indicates that the point lies farther from the mean vector in terms of Mahalanobis distance.

An explanation $E \subseteq V$ can be any subset from the power set $2^V$. The EMD algorithm generates a set of candidate explanations $E_{\text{set}}$ for a given (presumably anomalous) data point $z$, such that for each subset $\phi \in E_{\text{set}}$, the rank difference exceeds a predefined threshold of $\mu + c \cdot \sigma$, where $\mu$ and $\sigma$ are the mean and standard deviation of all rank differences. This ensures that the subset explains why $z$ is anomalous. The size of each candidate subset $\phi$ is restricted to at most $k_0$. If no such subset satisfies the condition, the algorithm returns an empty set.

We compute the belief of an explanation $\phi \in E_{\text{set}}$ using the standard deviation $\sigma$ of the rank difference between $R_V(x)$ and $R_{V \setminus A}(x)$ across all instances. Specifically, the belief is calculated as:

\[
Bel(z,\phi) = \frac{R_{V \setminus A}(z) - R_V(z)}{\sigma}
\]

This value represents how many standard deviations the rank difference for $z$ deviates from the mean of all rank differences. In other words, it measures the Mahalanobis distance of the rank difference for $z$ from the mean. Each candidate subset $\phi$ and its corresponding belief value is then provided as input to the Dempster-Shafer evidence combination method~\cite{Demp68}. The subset with the highest belief score is considered the most valid explanation.

\subsection{Explanation using iForest (EiForest)}

A well-known anomaly detection algorithm, iForest~\cite{liu2008isolation}, recursively partitions the data by randomly selecting a feature and its values for splitting. Data instances that become isolated in earlier splits are considered anomalies. We used iForest to construct a forest of $T$ such random partition trees. Let $P(z)$ denote the set of $T$ paths that lead to $z$. For a given instance $z$, we identify the set of features $V_P \subseteq V$ that appear on at least one path in $P(z)$, contributing to the isolation of $z$. For each variable $v \in V_P$, we construct an 8-dimensional summary feature vector $F_v^z$ using the paths leading to $z$ that contain $v$ (see Table~\ref{tab:iforest_summary} for details). We then build the set of summary vectors $F_v$ for all points and all variables in the dataset. Next, we compute the Mahalanobis distance $\pi(v)$ from the mean of $F_v$ for each $v \in V$. Finally, the top $k$ variables, sorted in decreasing order of $\pi(v)$, are selected as the explanation $E$.

\subsection{Explanation using Maximal Isolation (EMI)} 

We propose a method based on \emph{Integer Linear Programming} (ILP) that isolates an anomalous point to the maximum possible extent. The explanation $E$ generated by this method, called \emph{Explanation using Maximal Isolation} (EMI), is a conjunction of $L$ specified conditions. These conditions, when applied as filters on the entire dataset, minimize the number of points other than the anomalous point that satisfy all $L$ conditions. 

Given a set of features $V$ and an anomalous point $z$ to be explained, the explanation takes the form:
\[
\text{AND}(v (\leq \text{ or } \geq) z_v), \quad v \in E, \; E \subset V, \; |E| = L
\]
where $z_v$ is the value of $z$ for feature $v$. These $L$ conditions serve as an explanation for the anomalous nature of $z$ because they describe how $z$ differs from the rest of the dataset.

Table~\ref{tab:ILP} describes the ILP formulation in detail. Constraints $C_3$, $C_4$, $C_5$, and $C_6$ ensure that $\mathbf{y}[j]$ becomes $1$ if and only if the $j^{\text{th}}$ point violates at least one condition in the explanation. The objective function maximizes the number of such points, effectively minimizing the number of other points that satisfy all conditions along with $\mathbf{z}$, the anomalous point to be explained.

\subsection{Constructing the Silver Standard Ground Truth Explanations}

We have FS of 11,460 companies in our dataset, out of which 2727 were assigned the silver label {\small {\sf MISINFO}} using the method explained earlier. But what is the explanation (as a subset of our 42 financial variables) for each of these 2727 companies? Manually identifying such a subset involves too much efforts, time, and subjectivity. We now propose a novel automated method for constructing silver standard explanations for each of these 2727 companies. The idea is to use the sentences marked as {\small {\sf ADV\_REMARK}} in the associated audit report for the FS of a company in a particular year, extract financial variables mentioned in each of these sentences (labeled {\small {\sf ADV\_REMARK}}), and use that set as the explanation. However, there is a subtlety here: each financial variable mentioned in the sentence needs to be mapped to one of the features used in the structured representation of the FS. 

Example: in the sentence {\small {\tt In respect of disputed [Income tax] and [Sales Tax] cases , the company has not ascertained the amount of such dues .}}, the financial variables mentioned are enclosed in square brackets. But neither of these match with any feature in the  list of features.  In order to identify the related feature, we first assign 1 or more XBRL categories to the sentence using the techniques in~\cite{PAPV23} (explained below). XBRL is an international standard for the vocabulary to be used in financial reporting\footnote{www.xbrl.org}. We use the 1724 categories in the XBRL 2012 taxonomy as a set of standardized names of  financial variables. Then one of the XBRL categories assigned to this sentence by the method is {\small {\sf Taxes Payable Sales Tax}}; since this name is not among the 42 financial variables we use, we assign its parent XBRL category {\small {\sf Total Current Liabilities}} (which {\em is} present among the list of features) as one of the ground truth variables making this particular company anomalous. 

The method for assigning 0, 1 or more XBRL categories as the silver standard explanation for explaining why a particular company (i.e., its FS) is labeled {\small {\sf MISINFO}} in the database $D_{str}^{(SL)}$ is now explained~\cite{PAPV23}. This corresponds to the task B mentioned in \cite{PAPV23}. For year $Y$ and company $C$ which has the silver label {\small {\sf MISINFO}}, we take the sentences (there are at least two) from the associated audit report which are labeled as {\small {\sf ADV\_REMARK}} by both T5 and MISTRAL, as explained earlier. For each XBRL category $X$ having $|X| = k$ words in it, we consider windows of sizes $k-1$, $k$, $k+1$ and $k+2$ and obtain all possible contiguous word sequences in a given sentence $S$ by sliding these windows. The similarity of the category $X$ with the sentence is the highest Jaccard similarity with any of these subsets. While computing similarity, synonyms and morphological variations of a word are taken into consideration and stopwords are ignored. A category is matched with a sentence if the Jaccard similarity thus computed is above a threshold $\theta_1$ (we used $\theta_1$ = 0.6). Once the categories are assigned, we map them with the 42 financial variables used in this paper. In case we do not find an exact match for the XBRL category in these variables, we use the hierarchical relations in the XBRL taxonomy and consider its immediate next parent that will match to the variables (see the example above). This way, we take the $union$ of the variables assigned to the sentences as the ground truth of explanations for company $C$ and year $y$.

\subsection{Experiments and Results}

\textbf{Dataset}: We use the $D_{str}$ dataset of the specific year to generate explanations for any company $C$ (belongs to $D_{str}$) marked as suspicious. As mentioned in section \ref{sec:datasets}, for each year, the dataset consists of 2292 companies and values for 42 financial variables taken from Balance Sheet, Income Statement and Cash-flow Statement. 

\textbf{Parameter settings}: Parameter values for EMD algorithm are set as $c$ = 0.5 and $k_{0}$ = 2. For EiForest, we set $T$ = 1000 and retain top 5 features ($k$ = 5). For EMI, we set  the value of $L$ = 2. For SHAP and LIME we have retained top 5 features responsible for the prediction of the model.

\textbf{Baseline methods}: SHAP\cite{kuhn1953contributions} and LIME\cite{ribeiro2016should} are two widely used explanation generation methods for classification and regression. Both need an auxiliary classifier for generating the explanations for anomalous instances; we use Random Forest classifier for this purpose. We use the 2024 subset of the the silver label dataset $D_{str}^{SL}$ to train the Random Forest classifier. 

For illustrative purposes, we take the top 20 companies whose FS for 2014 were identified as anomalous by SVR M15 regression model for year 2014. We compare the explanations generated by SHAP, LIME and our methods (EMD, EiForest, EMI) for these companies and compare these explanations with the silver standard explanations (explained earlier); see Table~\ref{tab:exp_results}, which also includes the silver standard ground truth explanations. The Table includes only 7 of these companies (out of 20) because for the remaining 13 companies, either (i) the the intersection of sentences from audit reports classified by T5 and MISTRAL as ADV\_REMARK is empty (thus there is no misinformation in these FS as per our criterion); or (ii) the intersection of the generated explanations (by all 5 methods) with silver standard explanation is empty. We used precision $P$, recall $R$ and $F_{1}$ measure for each generated explanation as evaluation metrics, by comparing it with the  generated explanation using the silver-standard ground truth; see Table \ref{tab:exp_results_prf}. To cover such companies, We could relax our criterion, and generate explanations using sentences classified as adverse remarks by {\em either} T5 {\em or} MISTRAL. 

The EiForest algorithm exhibited improved results as compared to the original study~\cite{VSPP22}. In contrast, the other algorithms experience a decline in performance relative to their previously reported outcomes. The key factor contributing to this shift could be the increased number of financial variable (18 in~\cite{VSPP22}, current paper:42). EMI algorithm gives the average precision of 0.21, which is very near to the precision from EiForest. We have selected the top $k$ features for SHAP, LIME and EiForest algorithms. The value of $k$ can affect the precision values. The explanations given by SHAP and LIME are almost the same for most of the companies with just one variable difference between them. EMD algorithm could detect the variables correctly for 2 companies. The average $P, R, F_1$ values of EiForest algorithm are highest among all the algorithms. 

One limitation of our 3 methods for explanation generation is that they assume that the corresponding audit report is already available, which may not be the case in all possible scenarios. In case the audit report is not available, and so silver label ground truth is also not available, we can use SHAP or LIME to generate the explanation for a company whose FS is identified as anomalous.

\subsection{Generation of Suggestions for Auditors:}\label{sec:audit_suggestions}


For a sample company which was identified as anomalous in the experiments the explanation generation technique EiForest identified the following as the explanatory financial variables as to why FS is likely anomalous (corresponding XBRL category is also shown)\\*  
\begin{enumerate}
\item {\small {\tt \textbf{v40} $\rightarrow$ Net Cash Used In Investing Activities}} $\rightarrow$ {\small {\tt Net cash flow used in investing activities}}
\item {\small {\tt \textbf{v12} $\rightarrow$ Short Term Provisions}} $\rightarrow$ {\small {\tt Provisions}}
\item {\small {\tt \textbf{v15} $\rightarrow$ Fixed Assets}} $\rightarrow$ {\small {\tt Fixed Assets}}
\item {\small {\tt \textbf{v5}  $\rightarrow$ Deferred Tax Liabilities [Net]}} $\rightarrow$ {\small {\tt Net deferred tax liabilities}}
\item {\small {\tt \textbf{v28} $\rightarrow$ Total Assets}} $\rightarrow$ {\small {\tt Total Assets}}
\end{enumerate}

We mapped each {\small {\sf ADV\_REMARK}} sentence in the dataset $D_1$ to 0, 1 or more XBRL categories. Mapping the above XBRL categories corresponding to the above explanation, following are some of the matching adverse remarks in past audit reports:

\begin{enumerate}
\item {\small {\tt Adequacy of the provision made for meeting workers liability can not be ascertained}}
\item {\small {\tt Adequate provision for non recoverability has not been made for debtors}}
\item {\small {\tt No provision for diminution in the value of investment in shares of company X has been made.}}
\item {\small {\tt could not quantify the non provision of interest.}}
\item {\small {\tt No provision for pending sales tax amount of Rs. X has been made by the company.}}

\end{enumerate}

We could easily transform these selected adverse remarks (using LLMs or using simple linguistic rules for text transformation) as audit suggestions.
For example, the last two adverse remark sentences above can be transformed into audit suggestions as:
\begin{enumerate}
\small
    \item {\em Check if provisions for interests have been made.}
\item {\em Check if provisions for pending taxes have been made.}

\end{enumerate}


\begin{center}
\footnotesize
\begin{landscape}
\begin{longtable}[ht]{    |   p{0.03\textwidth}   |   p{0.18\textwidth}   |    p{0.25\textwidth}  |   p{0.2\textwidth}   |   p{0.18\textwidth}   |   p{0.18\textwidth} |   p{0.18\textwidth}   | p{0.18\textwidth} |}
\hline
\textbf{No.} &   \textbf{Company}    &   \textbf{Ground truth}   &   \textbf{SHAP}   &   \textbf{LIME}   &   \textbf{EMD}    &   \textbf{EiForest}   &   \textbf{EMI}\\\hline\hline
1	&	La Tim Metal	&	$v_{15}$, $v_{33}$, $v_{40}$	&	$v_{2}$, $v_{3}$, $v_{36}$, $v_{37}$,$v_{38}$	&	$v_{2}$, $v_{3}$, $v_{30}$, $v_{36}$, $v_{38}$	&	NA	&	$\mathbf{v_{40}}$, $v_{12}$, $\mathbf{v_{15}}$, $v_{5}$, $v_{28}$	&	$v_{26}$ $\leq$ 0.01 $\land$ $v_{41}$ $\geq$ 0.01 
\\\hline 
2	&	Unimers India	&	$v_{13}$,  $v_{21}$, $v_{27}$, $v_{36}$ , $v_{39}$, $v_{40}$, $v_{41}$	&	$v_{2}$, $v_{3}$, $\mathbf{v_{36}}$, $v_{37}$, $v_{38}$	&	$v_{2}$, $v_{3}$, $v_{11}$, $\mathbf{v_{36}}$, $v_{38}$	&	$v_{19}$, $v_{6}$, $v_{11}$, $v_{32}$	&	$v_{10}$, $v_{24}$, $v_{42}$, $v_{35}$, $v_{21}$	&	$\mathbf{v_{27}}$ $\leq$ 38.26 $\land$ $v_{19}$ $\geq$ 38.08 
\\\hline 
3	&	Nicco Uco Fin	&	$v_{3}$, $v_{13}$, $v_{15}$, $v_{16}$, $v_{17}$, $v_{27}$, $v_{34}$, $v_{39}$, $v_{41}$ 	&	$v_{2}$, $\mathbf{v_{3}}$, $v_{36}$, $v_{37}$, $v_{38}$	&	$v_{2}$, $\mathbf{v_{3}}$, $v_{7}$, $v_{36}$, $v_{38}$	&	$\mathbf{v_{13}}$, $v_{32}$, $v_{42}$, $v_{7}$	&	$v_{32}$, $\mathbf{v_{39}}$, $v_{7}$, $v_{19}$, $\mathbf{v_{17}}$	&	$v_{2}$ $\leq$ -564.2 $\land$ $v_{19}$ $\leq$ 8.7
\\\hline 
4	&	Entegra	&	$v_{40}$, $v_{41}$, $v_{9}$	&	$v_{2}$, $v_{11}$, $v_{36}$, $v_{37}$,$v_{38}$	&	$v_{2}$, $v_{3}$, $v_{11}$, $v_{36}$, $v_{38}$	&	$v_{32}$, $v_{39}$, $v_{11}$	&	$v_{11}$, $v_{1}$, $v_{7}$, $v_{34}$, $\mathbf{v_{41}}$	&	$v_{8}$ $\leq$ 0.08 $\land$ $v_{11}$ $\geq$ 162.68
\\\hline 
5	&	Quintegra Solut	&	$v_{2}$,$v_{40}$,$v_{41}$	&	$\mathbf{v_{2}}$, $v_{3}$, $v_{36}$, $v_{37}$,$v_{38}$	&	$v_{1}$, $\mathbf{v_{2}}$, $v_{3}$, $v_{36}$, $v_{38}$	&	NA	&	$v_{3}$, $v_{36}$, $v_{6}$, $v_{25}$, $v_{30}$	&	$v_{3}$ $\leq$ -108.29 $\land$ $\mathbf{v_{2}}$ $\geq$-135.11
\\\hline 
6	&	Modipon	&	 $v_{2}$, $v_{9}$, $v_{12}$, $v_{13}$, $v_{40}$, $v_{41}$ 	&	$\mathbf{v_{2}}$, $v_{3}$, $v_{30}$, $v_{37}$, $v_{38}$	&	$\mathbf{v_{2}}$, $v_{3}$ $v_{5}$, $v_{30}$, $v_{37}$	&	$\mathbf{v_{41}}$, $v_{6}$, $v_{32}$	&	$v_{1}$, $\mathbf{v_{41}}$, $v_{42}$	&	$v_{8}$ $\leq$ 21.7 $\land$ $v_{6}$ $\geq$ 21.69
\\\hline 
7	&	Encore Software	&	$v_{33}$, $v_{39}$, $v_{13}$	&	$v_{2}$, $v_{3}$, $v_{36}$, $v_{37}$,$v_{38}$	&	$v_{2}$, $v_{3}$, $v_{36}$, $v_{37}$,$v_{38}$	&	$v_{17}$, $v_{11}$, $v_{32}$	&	$v_{2}$, $\mathbf{v_{13}}$, $v_{11}$, $v_{7}$, $v_{27}$	&	$\mathbf{v_{13}}$ $\leq$ 24.6 $\land$ $v_{11}$ $\geq$ 23.27
\\\hline 
\caption{Explanations generated by all the methods.}
\label{tab:exp_results}
\end{longtable}

\begin{longtable}[ht]{ | p{0.04\textwidth} | p{0.22\textwidth} | p{0.045\textwidth} | p{0.045\textwidth} | p{0.045\textwidth} | p{0.045\textwidth} | p{0.045\textwidth} | p{0.045\textwidth} | p{0.045\textwidth} | p{0.045\textwidth} | p{0.045\textwidth} | p{0.045\textwidth} | p{0.045\textwidth} | p{0.045\textwidth} | p{0.045\textwidth} | p{0.045\textwidth} | p{0.045\textwidth} |}
    \hline
 &       &   \multicolumn{3}{p{0.05\textwidth}|}{\textbf{SHAP}}  &   \multicolumn{3}{p{0.05\textwidth}|}{\textbf{LIME}}  &  \multicolumn{3}{p{0.05\textwidth}|}{\textbf{EMD}}   &   \multicolumn{3}{p{0.05\textwidth}|}{\textbf{EiForest}}  &   \multicolumn{3}{p{0.05\textwidth}|}{\textbf{EMI}}    \\\hline
    \textbf{No.}    &    \textbf{Company}   &   \textit{P}  &   \textit{R}  &   \textit{F}  &   \textit{P}  &   \textit{R}  &   \textit{F}&    \textit{P}  &   \textit{R}  &   \textit{F} &   \textit{P}  &   \textit{R}  &   \textit{F}  &   \textit{P}  &   \textit{R}  &   \textit{F}  \\\hline  
1	&	La Tim Metal	&	0.00	&	0.00	&	0.00	&	0.00	&	0.00	&	0.00	&	0.00	&	0.00	&	0.00	&	0.40	&	0.67	&	0.50	&	0.00	&	0.00	&	0.00
\\\hline 
2	&	Unimers India	&	0.20	&	0.14	&	0.17	&	0.20	&	0.14	&	0.17	&	0.00	&	0.00	&	0.00	&	0.20	&	0.14	&	0.17	&	0.50	&	0.14	&	0.22
\\\hline 
3	&	Nicco Uco Fin	&	0.20	&	0.11	&	0.14	&	0.20	&	0.11	&	0.14	&	0.25	&	0.11	&	0.15	&	0.40	&	0.22	&	0.29	&	0.00	&	0.00	&	0.00
\\\hline 
4	&	Entegra	&	0.00	&	0.00	&	0.00	&	0.00	&	0.00	&	0.00	&	0.00	&	0.00	&	0.00	&	0.20	&	0.33	&	0.25	&	0.00	&	0.00	&	0.00
\\\hline 
5	&	Quintegra Solut	&	0.20	&	0.33	&	0.25	&	0.20	&	0.33	&	0.25	&	0.00	&	0.00	&	0.00	&	0.00	&	0.00	&	0.00	&	0.50	&	0.33	&	0.40
\\\hline 
6	&	Modipon	&	0.20	&	0.17	&	0.18	&	0.20	&	0.17	&	0.18	&	0.33	&	0.17	&	0.22	&	0.33	&	0.17	&	0.22	&	0.00	&	0.00	&	0.00
\\\hline 
7	&	Encore Software	&	0.00	&	0.00	&	0.00	&	0.00	&	0.00	&	0.00	&	0.00	&	0.00	&	0.00	&	0.20	&	0.33	&	0.25	&	0.50	&	0.33	&	0.40
\\\hline 
    &   \textbf{Average}    &   0.11	&	0.11	&	0.11	&	0.11	&	0.11	&	0.11	&	0.08	&	0.04	&	0.05	&	\textbf{0.25}	&	\textbf{0.27}	&	\textbf{0.24}	&	0.21	&	0.12	&	0.15
\\\hline
\caption{Precision, Recall and $F_{1}$ measure for all methods}
\label{tab:exp_results_prf}
\end{longtable}
\end{landscape}

\end{center}

\section{Conclusions and Future Work}\label{sec:conc}

Financial auditing plays an important role in maintaining the trust and transparency of the company. Detecting misinformation present in the Financial Statements is an important outcome expected from an audit. In this paper, we propose an integrated collection of AI-based techniques, based on previous work~\cite{SVAP22},~\cite{PAPV23},~\cite{VSPP22}, to provide assistance to human auditors, with the goals of reducing the time, efforts and subjectivity involved in the auditing process, as well as increasing its effectiveness. The idea is to provide effective assistance to human auditors by extracting knowledge from past corpus of FS and associated audit reports. 

We web-scraped gathered a broad range of financial variables from three types of FS. We proposed unsupervised techniques (unsupervised anomaly detection, as well as regression-based) for detecting misinformation in FS and showed that they they detect companies whose FS are likely containing  misinformation with good precision. Due to the absence of ground truth about FS having misinformation, we used NLP techniques and Language Models on audit reports to generate the silver-standard labels for this FS data. The techniques showed an improved precision on the much larger dataset used in this paper, as compared to the earlier papers. To identify precisely where the misinformation is present, we proposed techniques to generate explanations for the identified suspicious companies using the explainable AI (XAI) and other explainability techniques, as proposed in our previous work. Instead of manually validating the generated explanations (as was done earlier), we used the Jaccard Similarity technique and XBRL categories to get silver-standard ground truth for explanations. We assign the XBRL categories most similar to the audit report sentences which as adverse remarks, map the assigned XBRL categories to the financial variables used in this paper and consider the union of mapped variables as the silver standard ground truth. We have implemented the complete process for providing such assistance to the auditors, without involving any manual labour. This reduces the subjectivity involved in our previous work. Our techniques provide a focused area that seems suspicious, which could be helpful for the auditors while performing the audit.

For future work, we are working on integrating several other components of our research into a system that can be used by human auditors to get some assistance for improving the quality and effectiveness of the audits. Measuring the efficacy and usability of such a system, particularly the trustworthiness of the misinformation and its explanation is an important issue that needs to be tackled. Gathering auditor feedback about the detected misinformation, its explanation as well as generated audit suggestions and learning from the feedback to incrementally improve the systems are an important aspect that we will work on. We are also working on incorporating other sources of auditing knowledge, in the form of knowledge graphs, to improve the quality of generated outputs. 

\section{Statements and Declarations}

\subsection{Conflict of Interest Declaration}

On behalf of all authors, the corresponding author states that there is no conflict of interest.

\subsection{Competing Interests}
Not Applicable

\subsection{Funding Information}
We have not received any fundings.

\subsection{Author contribution}
•	Kshitij Madhav Jadhav – Conceptualization, Data curation, Formal analysis, Investigation, Project administration, Software, Writing – original draft, Visualization\\
•	Sudhodhan Vaishampayan- Conceptualization, Formal analysis, Investigation, Methodology, Writing – review and editing, Supervision\\
•	Manoj Apte- Conceptualization, Formal analysis, Methodology, Writing – review and editing, Supervision\\
•	Sachin Pawar- Conceptualization, Formal analysis, Methodology, Writing – review and editing, Supervision\\
•	Nitin Ramrakhiyani- Conceptualization, Formal analysis, Methodology, Writing – review and editing, Supervision\\
•	Girish Keshav Palshikar- Conceptualization, Methodology, Formal analysis, Project administration, Writing – original draft, Supervision

\subsection{Data Availability Statement}

Data used in the study is publicly available on \url{https://www.moneycontrol.com/}. Further processing on data is done as per the steps given in the manuscript.

\subsection{Research Involving Human and /or Animals}
Not Applicable

\subsection{Informed Consent}
Not Applicable

\subsection{Consent to Publish declaration}
Not Applicable

\subsection{Consent to Participate declaration}
Not Applicable

\subsection{Ethics declaration}
Not Applicable

\bibliography{sn-bibliography}

\end{document}